\documentclass[graybox]{article}

\usepackage{mathptmx}       % selects Times Roman as basic font
\usepackage{helvet}         % selects Helvetica as sans-serif font
\usepackage{courier}        % selects Courier as typewriter font
\usepackage{type1cm}        % activate if the above 3 fonts are
\usepackage{makeidx}         % allows index generation
\usepackage{graphicx}        % standard LaTeX graphics tool
\usepackage{multicol}        % used for the two-column index
\usepackage[bottom]{footmisc}% places footnotes at page bottom

\makeindex             % used for the subject index
\usepackage{amsmath}
\usepackage{amsfonts}
\usepackage{bigints}
\usepackage{amssymb}
\usepackage{mathtools}

\newcommand{\bx}{\mathbf{x}}
\newcommand{\EE}{\mathbb{E}}      % esperance
\newcommand{\unif}[2]{\mathcal{U}(#1,#2)}

\newcommand{\un}{\mathbf{1}}    % vecteur de 1   %\def \UN{\hbox{1}\!\hbox{I}}
\newcommand{\R}{\mathbb{R}}     % reels          %\def 

\begin{document}

\title{Sample selection from a given dataset to validate machine learning models}
% Use \titlerunning{Short Title} for an abbreviated version of
% your contribution title if the original one is too long}
\author{Bertrand Iooss\\
% Use \authorrunning{Short Title} for an abbreviated version of
% your contribution title if the original one is too long
EDF R\&D, 6 Quai Watier, 78401 Chatou, France\\
SINCLAIR AI Lab., Saclay, France}

% Use the package "url.sty" to avoid
% problems with special characters
% used in your e-mail or web address
%
\maketitle

\abstract{
The selection of a validation basis from a full dataset is often required in industrial use of supervised machine learning algorithm.
This validation basis will serve to realize an independent evaluation of the machine learning model. 
To select this basis, we propose to adopt a ``design of experiments'' point of view, by using statistical criteria. 
We show that the ``support points'' concept, based on Maximum Mean Discrepancy criteria, is particularly relevant. An industrial test case from the company EDF illustrates the practical interest of the methodology.}

%%%%%%%%%%%%%%%%%%%%%%%%%%%%%%%%%%%%%%%%%%%%%%
\section{Introduction}
\label{sec:1}

With the development of automatic diagnostics based on statistical predictive models, coming from any supervised machine learning (ML) algorithms, important issues about model validation have been raised. 
For example in the industrial non-destructive testing field (e.g. for aeronautic or nuclear industry), generalized automated inspection (that will allow large gain in terms of efficiency and economy) has to provide high guarantees in terms of performance. 
In this case, it is necessary to be able to select a validation data basis that will not be used for the training nor the selection of the ML model \cite{eni19,hawpat21}.
This validation data basis (also referred as verification data in the literature) has not to be communicated to the ML developers because it will serve to realize an independent evaluation of the provided ML model (applying a cross validation method is then not possible).
This validation sample is typically used to provide prediction residuals (which can be finely analyzed), as well as average ML model quality measures (as the mean square error in a regression problem or the misclassification rate in a classification problem).

In this paper, we address the particular question about the way to select a ``good'' validation basis from a dataset useful to specify a ML model. 
We use indifferently the term ``validation'' and ``test'' for the basis (also called sample) because we restrict our problem to the distinction between a learning sample (which includes the ML fitting and selection phases) and a test sample.
%Basic and simplistic elements about this problem are given in every ML handbooks (see, e.g., \cite{hastib09}).
An important question is the number and the location of these test points.
For the size of the test sample, no general theoretical rule can be given while the classical ML handbooks \cite{hastib09,gooben16} provide different heuristic rules (as, e.g., $80\% / 20\%$ between the learning and test samples and $50\% / 25\% / 25\%$ between the learning, model selection and test samples).

In our validation basis selection problem, the dataset already exists so the problem turns to selecting a certain number of points in a finite collection of points. 
For simplicity, our work is limited to a supervised classification problem with two clusters: a validation sample is extracted in each sub-dataset (corresponding to each cluster). 
%This approach will be easily generalized to any number of clusters.
For the test sample location issue, simple selection algorithms are sometimes insufficient to ensure the representativity of the validation basis, in particular for small-size and highly unbalanced datasets.
Indeed, the simplest and usual practice to build a test sample is to randomly extract an independent Monte Carlo sample \cite{hastib09}.
If the sample size is small, as for the space-filling design issues \cite{fanli06}, it is well known that the proposed points can be badly localized (test samples too close from learning points or leaving large input space subdomain unsampled).
Therefore, a supervised selection based on statistical criteria is necessary. 
%In large dataset cases, an appropriate selection will also lead to more robust ML model predictive capabilities evaluations than a random choice. 

A review of classical methods for solving this issue is given in \cite{borjir12}.
For example, CADEX \cite{kensto69} is a sequential selection algorithm of points inside a database to put in a validation basis, via inter-points distance computations.
From chemometrics, \cite{daswal02} complements this literature with cluster-based selection methods.
Several ideas have also been recently introduced in order to help interpreting the ML models \cite{mol19}.
It consists in identifying (in the dataset) the so-called prototypes (data instance representative of all the data) and criticisms (data instance not well represented by the set of prototypes).
To extract prototypes and criticisms, \cite{mol19} explains the principle of a greedy algorithm based on the Maximum Mean Discrepancy (MMD, see \cite{smogre07}).

Our work hybridizes the latter approach with the concepts of support points recently introduced by \cite{makjos18}, and which can be used to provide a representative sample of a desired distribution, or a representative reduction of a big dataset.
In Section \ref{sec:2}, the support points based algorithm is presented, with a simple application case.
Section \ref{sec:3} illustrates the practical interest of the methodology on an industrial test case.
Section \ref{sec:4} concludes with some perspectives of this work.

%%%%%%%%%%%%%%%%%%%%%%%%%%%%%%%%%%%%%%%%%%%%%%
\section{Use of support points}
\label{sec:2}

In this section, we use the recent work of \cite{makjos18} about a method to compact a continuous probability distribution $F$ into a set of representative points, called support points.
With respect to more heuristic methods for solving this problem, support points have theoretical guarantees in terms of the asymptotic convergence of their empirical distribution to $F$.
Moreover, the extraction algorithm is efficient in terms of computational cost, even for large-size test sample $N$ (up to $N=10^4$) and in high input space dimension $d$ (as large as $d=500$).

The construction of the support points is based on the optimization of the energy distance which is a particular case of the MMD criterion \cite{szeriz13}.
The MMD provides a distance between $F$ and a uniform distribution (via a kernel metric) and can be used with a relative good computational efficiency in high dimension (thanks to the kernel trick).
Let denote $\bx=(x_1,\ldots,x_d) \in \R^d$.
The discrete distribution of $N_v$ support points $\bx^{N_v}=(\bx^{(i)})_{i=1\ldots N_v}$ is denoted $F_{N_v}$ and the energy distance between $F$ and $F_{N_v}$ writes:
\begin{equation}\label{eq:energySP}
d_E^2(F,F_{N_v}) = \frac{2}{N_v} \sum_{i=1}^{N_v} \EE \|\bx^{(i)}-\zeta\| - \frac{1}{N_v^2} \sum_{i=1}^{N_v}\sum_{j=1}^{N_v} \EE \|\bx^{(i)}-\bx^{(j)}\| -  \EE \|\zeta-\zeta'\| 
\end{equation}
with $\zeta,\zeta'\sim F$ and by using the Euclidean norm.
The energy distance is always non-negative and equals zero if the two distributions are the same.
The support points $(\xi^{(i)})_{i=1\ldots N_v}$ are then defined by minimizing $d_E^2(F,F_{N_v})$.
Finding the support points corresponds to solving an optimization problem of large complexity, where $F$ is empirically known by the sample points (the dataset).
\cite{makjos18} provides an efficient algorithm to solve it.
The objective function is approximated by a Monte Carlo estimate, giving
\begin{equation}\label{eq:SPestim}
(\xi^{(i)})_{i=1\ldots N_v} = \arg\!\!\!\min_{\bx^{(1)},\ldots,\bx^{(N_v)}} \left( \frac{2}{N_v n} \sum_{i=1}^{N_v} \sum_{k=1}^n \|\bx^{(i)}-\bx'^{(k)}\| - \frac{1}{n^2} \sum_{i=1}^{N_v}\sum_{j=1}^{N_v} \EE \|\bx^{(i)}-\bx^{(j)}\| \right)
\end{equation}
where $(\bx'^{(k)})_{k=1\ldots n}$ is the $n$-size sample from $F$.
This cost function can be written as a difference of convex functions in $\bx^{N_v}$ and then can be minimized thanks to a formulation as a difference-of-convex program.
This procedure being quite slow, a combination of the convex-concave procedure (CCP) with resampling is used (see \cite{makjos18} and references therein for details) in order to obtain an efficient algorithm.
The examples given by \cite{makjos18} clearly show that support points distribution are more uniform than the ones of Monte Carlo and quasi-Monte Carlo samples \cite{fanli06}.

In the CCP procedure, the selected points are not extracted from the dataset but are the ``best'' points representative of the full dataset distribution.
Therefore, for our points selection problem, an additional step is required in order to find the $N_v$ representative points inside the dataset.
For each support point, we select the nearest dataset point and call this new algorithm SPNN (``support points nearest neighbor'').

\begin{figure}[!ht]
\begin{center}
   \includegraphics[width=\textwidth]{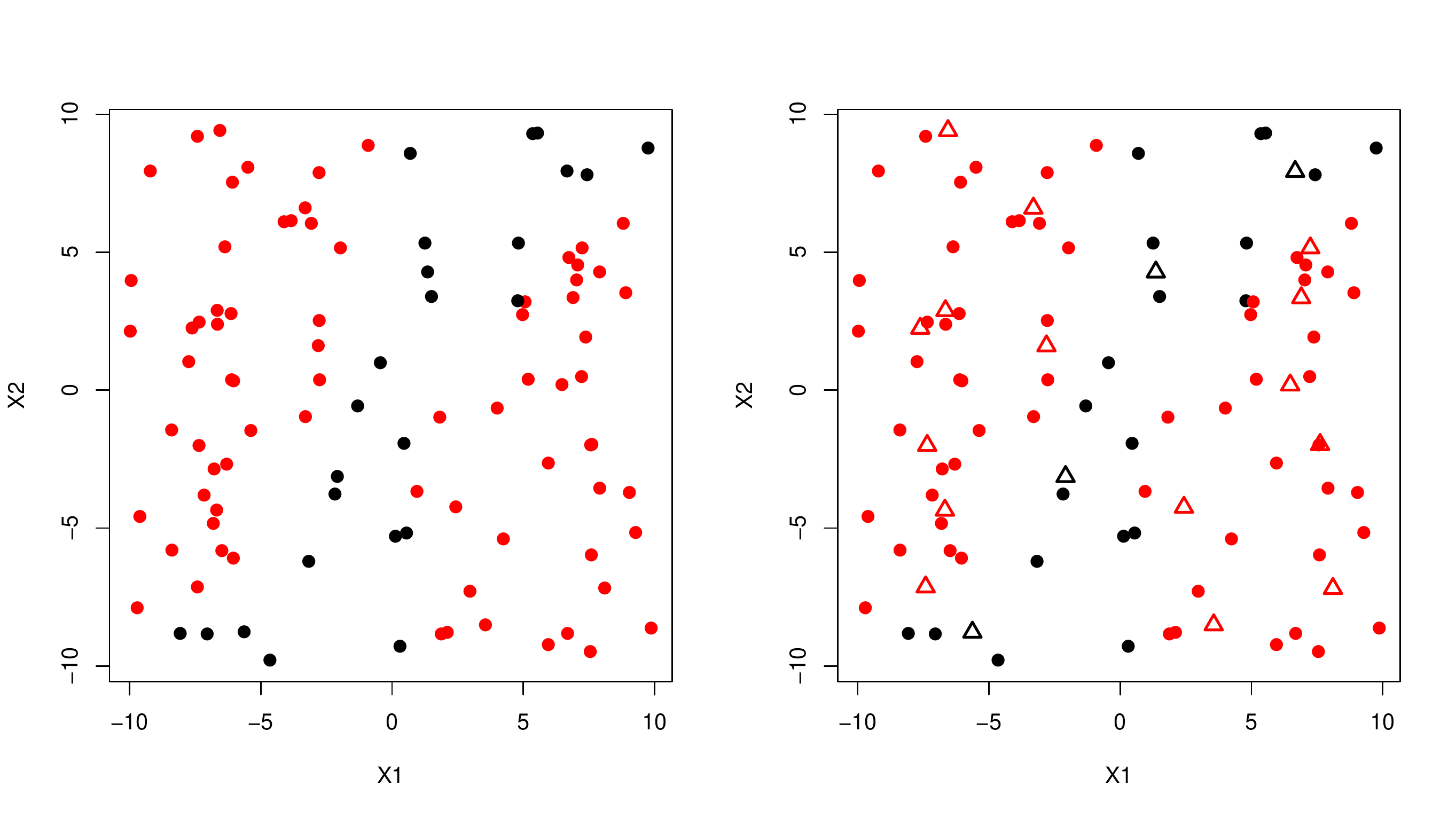} 
\end{center}

\caption{Indicator function example. Left: dataset points corresponding to the two clusters (black: $Y=0$, red: $Y=1$). Right: test points selected by the SPNN algorithm (triangle symbol).}
    \label{fig:SPNN2Dclass}
  \end{figure}

To illustrate SPNN on a toy example, we build a two-class two-dimensional ($d=2$) dataset of size $N=100$.
The classification model is the following: 
\begin{equation}
Y = \un_{X_1^2 - X_1 X_2 - X_1 - 3 > 0}
\end{equation}
with $\un_{(.)}$ the indicator function, $X_1 \sim \unif{-10}{10}$ and $X_2 \sim \unif{-10}{10}$.
The goal is to extract $20\%$ of points for the test sample, respecting the proportion of points in each class.
Applying the SPNN algorithm on the two sub-datasets (corresponding to each class) gives Fig. \ref{fig:SPNN2Dclass} which shows that the test points distribution in each class is quite satisfactory.

%%%%%%%%%%%%%%%%%%%%%%%%%%%%%%%%%%%%%%%%%%%%%%
\section{Application on an industrial use-case}
\label{sec:3}

This industrial problem aims at studying the fission products released in the primary circuit's water of the EDF nuclear reactors, during the load drop phase of the reactor cold shutdown. 
The available full dataset allows for $N=90$ observations containing $d=25$ covariates (describing the operation conditions of the reactor just before the shutdown) and the iodine activity level \cite{remdau18}.
The goal is to model the event that this iodine activity level exceeds a specific threshold, that can have large impact on the scheduled planning, so on operational costs.
Our classification dataset is well balanced as $48.9\%$ of the observations (called ``positive'') are above the threshold and $51.1\%$ of the observations (called ``negative'') are below the threshold.

For simplicity and, as it is not the subject of this work, we consider a naive logistic linear regression model (which predicts the probability for an individual to be positive) as the ML model (the probability threshold value of $0.5$ is used to assign each individual to one of the two classes).
To measure the quality of this ML model, we use the two main classification metrics: the error rate $\varepsilon$ (number of missclassified observations on total number of observations) and the sensitivity $\tau$ (number of well classified positive on the total number of positive observations).
Due to the large number of covariates relatively to the observations number, the ML model applied on the full dataset (or on any sub-sample) gives unsurprisingly an error rate of zero and a sensitivity of $100\%$.
By using a leave-one-out (LOO) procedure \cite{hastib09}, we are able to evaluate these metrics in prediction: $\varepsilon=18\%$ and $\tau=82\%$.
This LOO procedure is also used in the following tests on each learning sample (resulting from the extraction of the validation sample from the full sample). 

\begin{figure}[!ht]
\begin{center}
   \includegraphics[width=\textwidth]{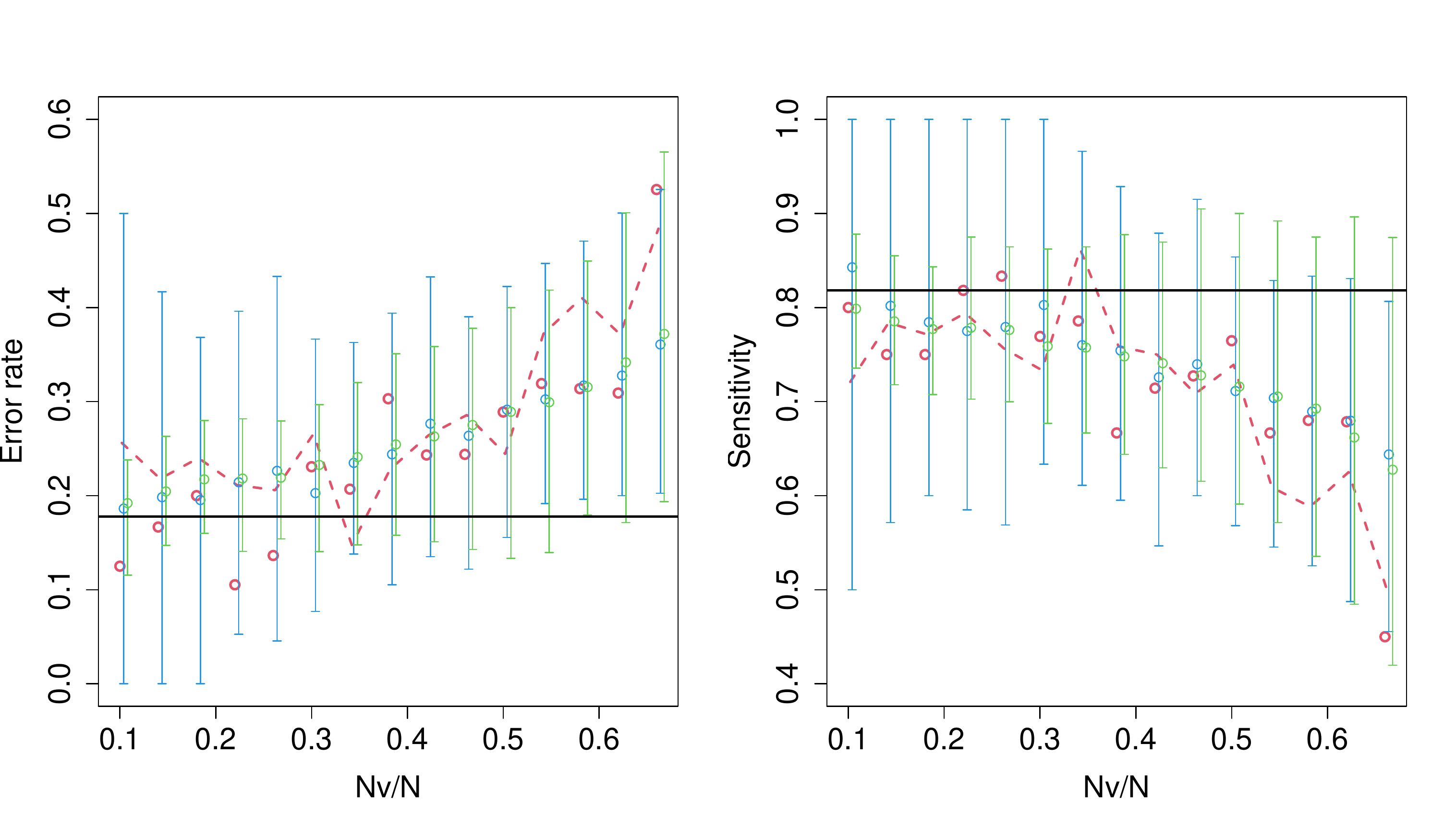} 
\end{center}

\caption{Classification metrics (error rate $\varepsilon$ at left and sensitivity $\tau$ at right) on the fission products dataset. Black line: reference values (LOO on full dataset). Red points (resp. dotted line): values from SPNN-based validation sample (resp. LOO-learning sample). Blue (resp. green) CI: $95\%$-CI from random validation samples (resp. LOO-learning samples).}
    \label{fig:SPNNPF}
  \end{figure}

Our goal is to study the capabilities of the SPNN algorithm in evaluating these metrics for different sizes of the validation sample (between $10\%$ and $66\%$ of the full sample size).
Figure \ref{fig:SPNNPF} provides the results that are compared to those obtained from a random sampling strategy.
Error rates and sensitivities seem adequately predicted from the SPNN-based validation samples, from ratio $N_v/N$ between $0.1$ and $0.35$.
Of course, this result is specific to our small-size use-case.
For such studies, the results also clearly show the inadequacy of the random validation samples to predict the ML model predictive capabilities.
Indeed, their confidence-intervals (CI) are huge and far from reference values.

%%%%%%%%%%%%%%%%%%%%%%%%%%%%%%%%%%%%%%%%%%
\section{Conclusion}
\label{sec:4}

In this work, the SPNN algorithm has been proposed for the selection of a test sample representative of a dataset.
It is not restricted to an hypercubic domain (no need to transform each input to $\unif{0}{1}$) as the classical space-filling criteria in the computer experiments literature \cite{fanli06}.
Moreover, compared to classical algorithms (as CADEX \cite{kensto69}), its computational cost does not depend on the dataset size and the data dimension.
Its main practical limitation is that it becomes prohibitive for a test sample size $N_v$ too large ($>10^4$).

Further improvements of this work would be interesting to study in a near future.
First, the approach gives equal importance to all the $d$ inputs.
It seems however useless to consider the inputs whose influence is negligible on the output.
A preliminary step would be useful to identify important inputs and to apply the test sample selection algorithm only on these components.
Second, new ideas for the support points definition can be developped, as for instance the use of the kernel Wasserstein distance \cite{ohpou20} instead of the energy distance.
Finally, this algorithm will also be useful for more complex classification problems where the inputs are temporal signals or images.
Specific kernels on the input space should be adapted to these cases.

%%%%%%%%%%%%%%%%%%%%%%%%%%%%%%%%
\section*{Acknowledgments}
This work has been funded by the international ANR project INDEX (ANR-18-CE91-0007) devoted to researches on incremental design of experiments.
The author is grateful to Emmanuel Remy, S\'ebastien da Veiga, Luc Pronzato and Werner M\"uller for giving ideas during this work, as well as Emilie Dautr\^eme, Vanessa Verg\`es and Marouane Il Idrissi for their help on the EDF dataset.
%Thanks to Grazia Vicario for the invitation to communicate this work.

%%%%%%%%%%%%%%%%%%%%%%%%%%%

%\input{referenc}

\end{document}